\documentclass[conference]{IEEEtran}
\IEEEoverridecommandlockouts
\usepackage{cite}
\usepackage{amsmath,amssymb,amsfonts}
\usepackage{algorithmic}
\usepackage{graphicx}
\usepackage{textcomp}
\usepackage{xcolor}
\usepackage{subfig}
\usepackage{balance}
\usepackage{colortbl}
\usepackage{booktabs}
\usepackage{hyperref}

\newcommand{\etal}{\emph{et al.}~}
\newcommand{\ie}{\emph{i.e.}~}

\newcommand{\etc}{\emph{etc}}

\def\BibTeX{{\rm B\kern-.05em{\sc i\kern-.025em b}\kern-.08em
    T\kern-.1667em\lower.7ex\hbox{E}\kern-.125emX}}
\begin{document}

\title{Convolutional Prompting for Broad-Domain Retinal Vessel Segmentation
\thanks{This work was supported by NSFC (62172420) and National High Level Hospital Clinical Research Funding (2022-PUMCH-C-61).}
\thanks{$^{*}$Corresponding author: Xirong Li (xirong@ruc.edu.cn)}
}

\author{
Qijie Wei$^{1}$ \qquad Weihong Yu$^{2,3}$ \qquad Xirong Li$^{1,3*}$ 
\\

$^{1}$Renmin University of China, China \\
$^{2}$Peking Union Medical College Hospital, China \\
$^{3}$Beijing Key Laboratory of Fundus Diseases Intelligent Diagnosis \& Drug/Device Development and Translation

}

\maketitle

\begin{abstract}
Previous research on retinal vessel segmentation is targeted at a specific image domain, mostly color fundus photography (CFP). In this paper we make a brave attempt to attack a more challenging task of broad-domain retinal vessel segmentation (BD-RVS), which is to develop a \emph{unified} model applicable to varied domains including CFP, SLO, UWF, OCTA and FFA. To that end, we propose \emph{Dual Convoltuional Prompting} (\texttt{DCP}) that learns to extract domain-specific features by localized prompting along both position and channel dimensions. \texttt{DCP} is designed as a plug-in module that can effectively turn a R2AU-Net based vessel segmentation network to a unified model, yet without the need of modifying its network structure. For evaluation we build a broad-domain set using five public domain-specific datasets including ROSSA, FIVES, IOSTAR, PRIME-FP20 and VAMPIRE. In order to benchmark BD-RVS on the broad-domain dataset, we re-purpose a number of existing methods originally developed in other contexts, producing eight baseline methods in total. Extensive experiments show the the proposed  method compares favorably against the baselines for BD-RVS.
\end{abstract}


\section{Introduction}

Retinal blood vessel characteristics are associated to both ocular and systemic health conditions \cite{fraz2014automated,li2021vessel,su2019exploiting}. For an accurate analysis of these vascular features, precise retinal vessel segmentation (RVS) in \emph{en face} retinal images is crucial. Moreover, the complexity of our retina requires varied fundus imaging techniques including color fundus photography (CFP), scanning laser ophthalmoscopy (SLO), ultra-widefield  fundus imaging (UWF), optical coherence tomography angiography (OCTA), fundus fluorescein angiography (FFA), \etc.  Images produced by a specific imaging technique form a specific \emph{domain}. For better visualization and diagnosis, cross-domain image registration and fusion are needed, for which RVS is often a prerequisite \cite{zhang2021two}. In such a context, an RVS model that universally works for varied domains will be handy.

\begin{figure}[!tb]
        \centerline{\subfloat[Narrow-domain RVS \label{fig:single-rvs}]{
            \includegraphics[width=0.7\columnwidth]{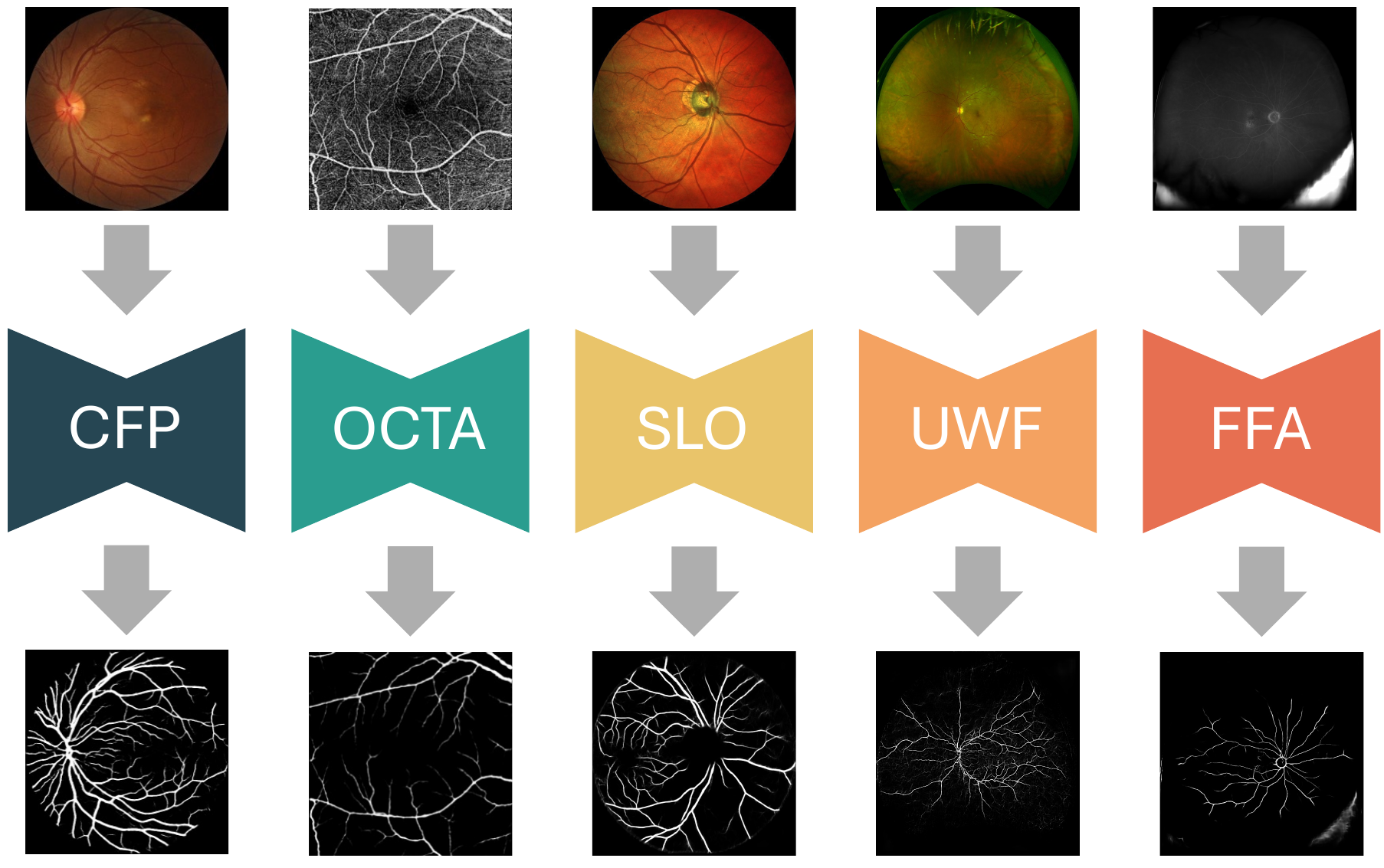}}}
             
        \centerline{\subfloat[Broad-domain RVS \label{fig:broad-rvs}]{
            \includegraphics[width=0.7\columnwidth]{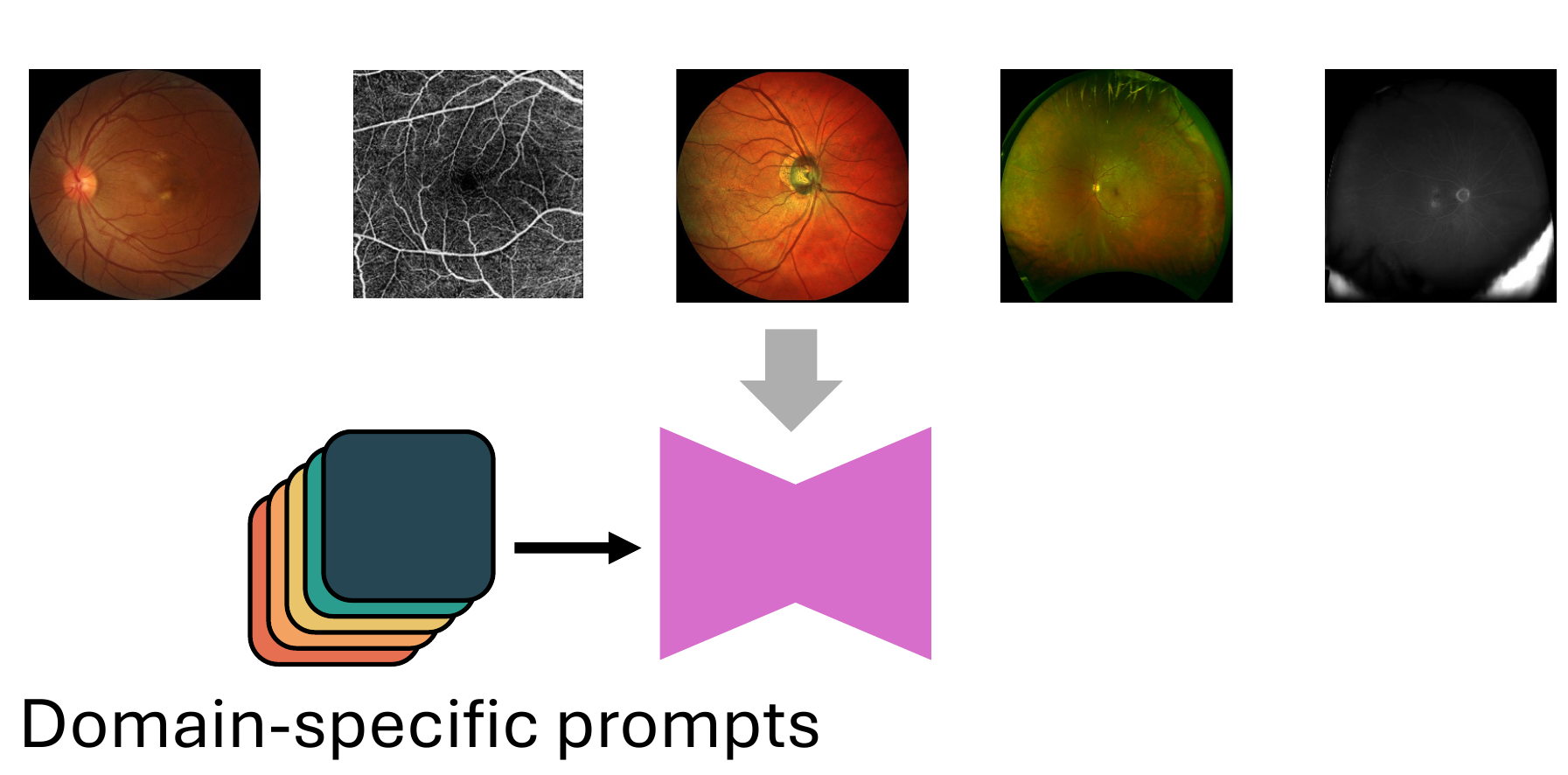}}}            
    \caption{\textbf{Two paradigms for retinal vessel segmentation (RVS)}: (a) narrow-domain and (b) broad-domain. This paper aims for the latter.}

    \label{fig:illu}
\end{figure}

Narrow-domain RVS has been extensively studied, mostly on the CFP domain, 
with improvements in network architectures\cite{lin2023stimulus,wang2022DANet,r2attunet}, training strategies\cite{iternet,tan2022retinal,xu2022local,sgl}, and both\cite{frunet}.
Furthermore, there are few works on other domains such as UWF \cite{pellegrini2017graph,prime-fp20} and OCTA \cite{octa-frnet,hu2021life}. 
Different from these efforts, in this paper we consider a more challenging task of \emph{broad-domain} retinal vessel segmentation (BD-RVS). As shown in Fig. \ref{fig:illu}, we aim for a \emph{unified} RVS model applicable to five distinct domains.

Developing a model for BD-RVS is nontrivial. Due to the large disparity in their visual appearance, simply training on domain-mixed data is ineffective for learning domain-specific features. Domain adaptation improves a model's performance on a target domain, yet practically at the cost of performance degeneration on the source domain where the model is originally trained \cite{ude}.
Prompt learning, originally developed for adapting pre-trained large language models to varied NLP tasks with minimal training data, is gaining popularity for both generic \cite{eccv22vpt,nips23cvp} and medical \cite{promptda} image analysis. For natural image classification,  Tsai \etal \cite{nips23cvp} introduce Convolutional Visual Prompt (\texttt{CVP}), which uses a single conv. layer as a prompt applied to the input image. Such a shallow prompt is unlikely to be sufficient to cover the large inter-domain divergence in retinal images. In the context of brain tumor segmentation, Lin \etal \cite{promptda} propose \texttt{prompt-DA}, which uses an extra domain classification network to extract domain-related features as a prompt and fuses the prompt with domain-independent features  in a holistic manner. Such a fusion strategy might fail to extract local domain-specific features important for segmenting thin vessels and capillaries. Hence,  although good performance has been reported on their own tasks, we argue that the current prompt learning methods are suboptimal for BD-RVS.
\begin{figure}[!tb]
   \centerline{
        \includegraphics[width=0.95\columnwidth]{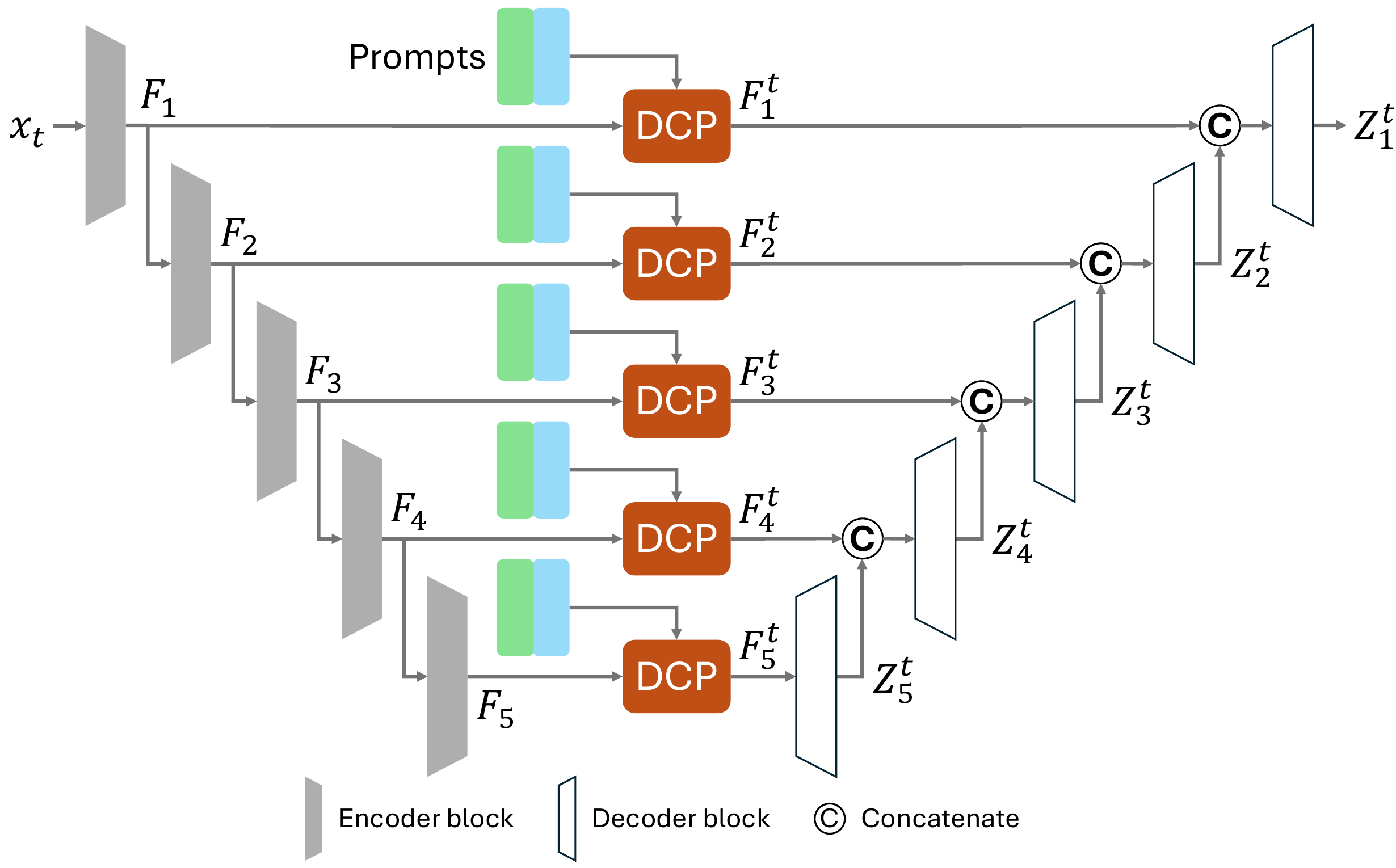}}
     \caption{\textbf{Our \texttt{DCP} method for broad-domain RVS }}
     \label{fig:overview}
\end{figure}

The main contributions of this paper are as follows: \\
$\bullet$ To the best of our knowledge, we are the first to attack the challenging task of BD-RVS, developing a unified model that works for five retinal-image domains including CFP, OCTA, SLO, UWF and FFA. \\
$\bullet$ We propose \emph{dual convolutional prompting} (\texttt{DCP}), extracting domain-specific features by localized prompting in both position and channel dimensions. \texttt{DCP} is designed as a plug-in module such that it can be used with a well establish RVS network without changing the network structure. \\
$\bullet$ Experiments on a broad-domain set, comprised of five public datasets,  show the viability of \texttt{DCP} for BD-RVS.  Code is available at \href{https://github.com/ruc-aimc-lab/dcp}{https://github.com/ruc-aimc-lab/dcp}. 


\section{Proposed Method}

\begin{figure*}[htb!]
    \centerline{
    \includegraphics[width=0.9\textwidth]{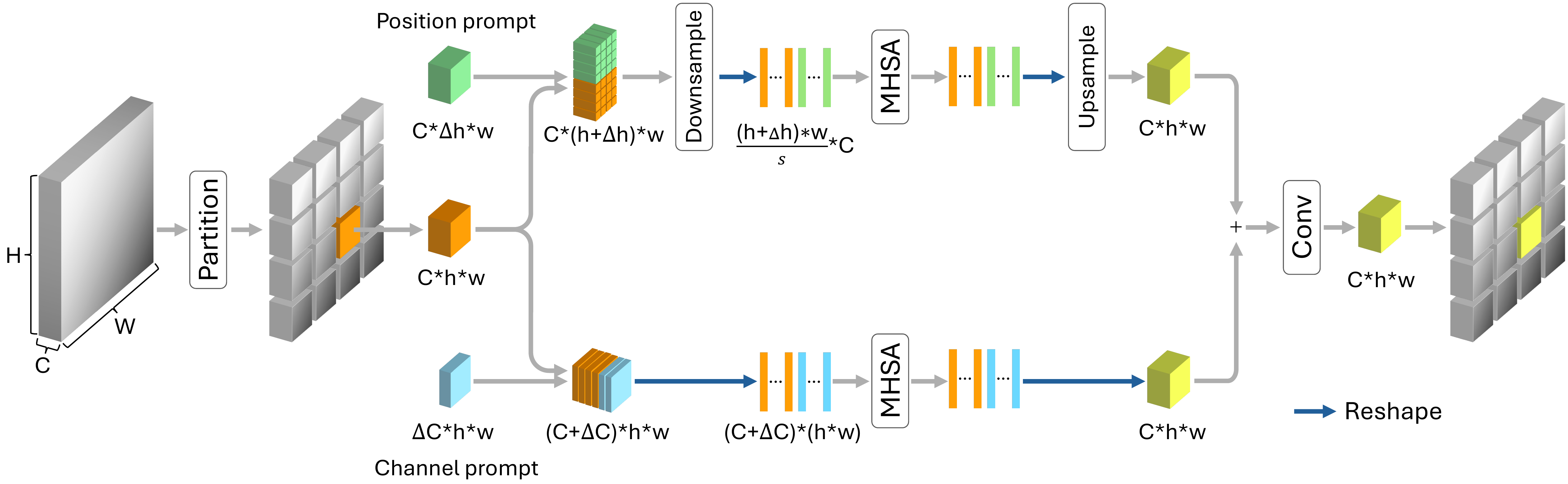}}
    \caption{\textbf{Proposed dual convolutional prompting (\texttt{DCP}) module}. Its input is the output feature map of a specific encoder of R2AU-Net. Based on the domain identity of the input image, \texttt{DCP} takes two domain-specific prompt tensors, which interact with the feature maps along the position and channel dimensions, respectively. For 
    \emph{localized} prompting, the feature map is partitioned into smaller (orange) patches. Once trained, the prompts are fixed. Best viewed in color.}
    \label{fig:dual}
\end{figure*}

We formalize the BD-RVS task as follows. We assume the availability of labeled training data from $T$ image domains. For each domain $t \in \{1, 2, ... , T\}$, we have a set of $n_t$ annotated images $\mathcal{N}_t=\{(x_{t}, y_{t}) \}$, where $x_{t}$ indicates a specific image with $y_t$ as its binary vessel mask. The goal of BD-RVS is to train a unified retinal vessel segmentation model based on the multi-domain training data $\mathcal{N}_1 \cup \ldots \mathcal{N}_T$.
Our method for BD-RVS is to inject domain-specific knowledge and consequently extract domain-specific features via dual convolutional prompting (\texttt{DCP}) into a well established network. In particular, our method works as a plug-in module so that the existing network needs no structural change. In what follows, we describe briefly the overall network in Sec. \ref{ssec:network} followed by \texttt{DCP} in Sec. \ref{ssec:dcp}.



\subsection{The Segmentation Network} \label{ssec:network}

We adopt R2AU-Net \cite{r2attunet} for its good performance on retinal vessel segmentation. R2AU-Net improves the classical U-Net network \cite{unet} by adding attention-enhanced recurrent residual blocks \cite{attunet,runet}. A common implementation of U-Net (and its variants like R2U-Net \cite{runet} and R2AU-Net) uses five convolutional encoders to reduce the input image into an array of progressively downsized feature maps. The feature maps then one-by-one go through five convolutional decoders to generate upsized feature maps. More formally, letting $F_i$ be the output of the $i$-th encoder and $Z_i$ the output of the $i$-th decoder, $i=1,\ldots,5$, the workflow of U-Net can be expressed as 
\begin{equation} \label{eq:unet}
\left\{
\begin{array}{ll}
F_i & \leftarrow Encoder_i(F_{i-1}), \\
Z_i & \leftarrow Decoder_i([Z_{i+1}; F_{i}]),\\
\end{array}
\right.
\end{equation}
where $F_0$ is the input image, $Z_6$ is null, and $Z_1$ is a probabilistic segmentation output. The input of each decoder is enhanced by recycling the output of the encoder at the same level through a skip connection. As illustrated in Fig. \ref{fig:overview}, Our \texttt{DCP} module is added to the skip connection, producing a domain-specific feature map $F^t_i$ that has the same shape as $F_i$. As such, the network structure of R2AU-Net requires no change. Feeding $F^t_i$ into the decoder yields $Z^t_i$. By executing $Decoder([Z^t_{i+1}; F^t_i])$, domain-specific decoding is achieved with ease.




\subsection{The Dual Convolutional Prompting Module} \label{ssec:dcp}

In order to extract domain-specific features, we design \texttt{DCP} as follows. Recall that each encoder of the segmentation network outputs an array of feature maps $F$. Suppose $F$ has $C$ channels, each with a spatial resolution of $H \times W$. Different channels typically capture different visual patterns, whilst feature values at a given spatial position indicate the local presence or absence of the patterns. Hence, jointly prompting along the position and channel dimensions is necessary. Moreover, we are inspired by translation invariance in convolutional neural networks, where the same conv. filter is applied to different positions of a given image or feature map. We thus choose to apply domain-specific prompts in a similar manner. Putting the above thoughts into practice, \texttt{DCP} has three blocks, \ie feature partition, dual prompting, and prompted-feature fusion, see Fig.~\ref{fig:dual}.

\textbf{Feature Partition}. 
For localized prompting, the feature maps $F$ are partitioned into $8\times 8$ non-overlapped patches, each sized to $C \times h \times w$, $h=\frac{H}{8}$, $w=\frac{W}{8}$. Since the spatial resolution of $F$ varies with the encoders, fixing the number of patches to 64 is computationally convenient. The same prompt tensors will be used for the individual patches.


\textbf{Position-wise Prompting}.
Each patch is concatenated with a domain-specific $C \times \Delta h \times w$ prompt along its height dimension ($\Delta h = h$ in this work). For position-wise feature interaction, we shall view the combined features as a sequence of $(h+\Delta h)\times w$ tokens. In order to make Transformer-based feature interaction computationally affordable, we reduce the sequence length by a scale factor of $s$, achieved by a conv. operation with stride  $\sqrt{s}$. For the five \texttt{DCP} modules from top to bottom shown in Fig. \ref{fig:overview}, $s$ is set to 64, 16, 4, 1, 1, respectively.
The shortened sequence of length $\frac{h+ \Delta h}{s}$ is then fed into a standard multi-head self attention (MHSA) block \cite{transformer}. The output sequence, with the prompt tokens removed, goes back to the original-patch size, by reshaping and bilinear-interpolation  upsampling.

\textbf{Channel-wise Prompting}. 
Each patch is concatenated with a domain-specific $\Delta C \times  h \times w$ prompt along its channel dimension (here $\Delta C = \frac{C}{4}$). For channel-wise feature interaction, we treat the combined features as a sequence of $C+\Delta C$ tokens and feed the sequence into an MHSA block. The output sequence, with the prompt tokens removed, is reshaped back to the original-patch size. 

\begin{figure}[htb!]
    \centerline{
    \includegraphics[width=0.49\textwidth]{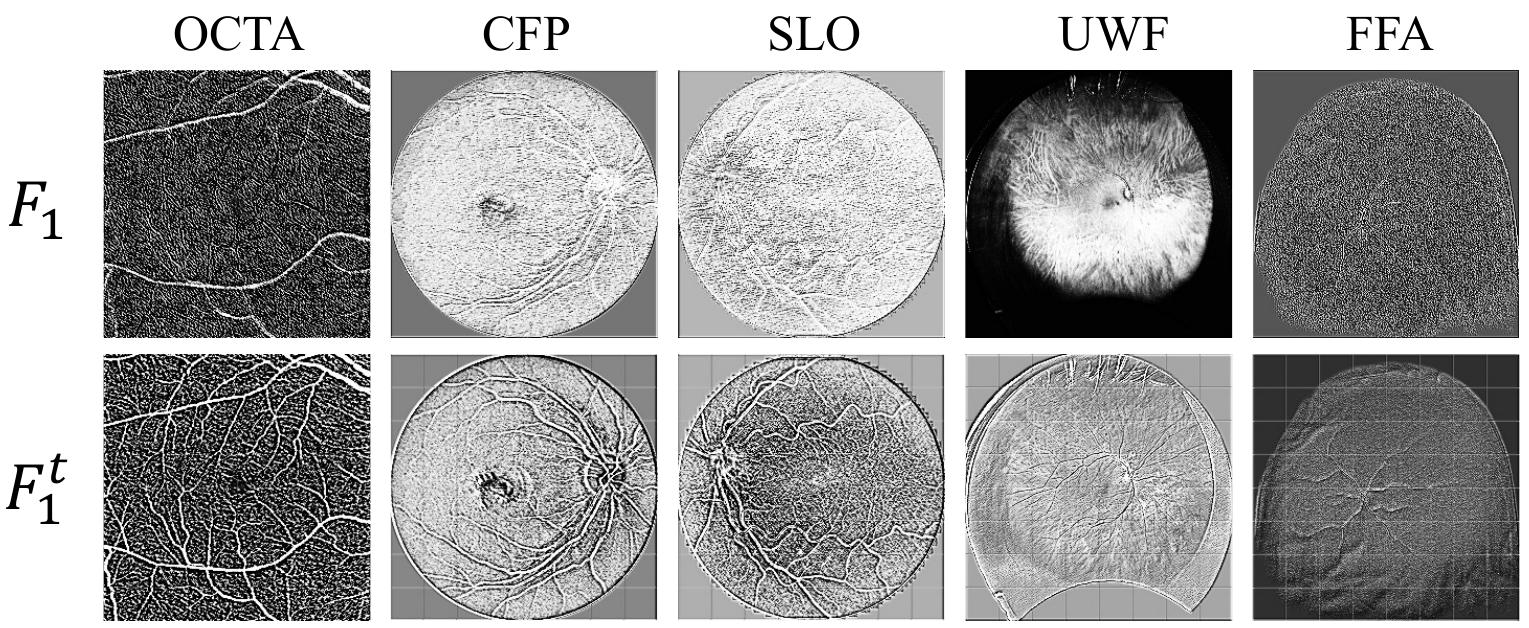}}
    \caption{\textbf{Visualization of the input ($F_1$) and output ($F^t_1$) of \texttt{DCP}}. For all the five modalities, vessel-related patterns are noticeably enhanced.}
    \label{fig:visualization}
\end{figure}

\textbf{Prompted-feature Fusion}. 
As shown in Fig. \ref{fig:dual}, for fusing the dually prompted patches, 
we simply apply element-wise addition followed by a conv. layer. See Fig. \ref{fig:visualization} for the effect of \texttt{DCP} on the feature maps.

\section{Experiments}
\subsection{Experimental Setup}

\textbf{Datasets}.
To build a broad-domain dataset, 
we adopt five public datasets, one per modality. That is, ROSSA\footnote{\href{https://github.com/nhjydywd/OCTA-FRNet}{https://github.com/nhjydywd/OCTA-FRNet}} with 918 OCTA images \cite{octa-frnet}, FIVES\footnote{\href{https://figshare.com/articles/figure/FIVES_A_Fundus_Image_Dataset_for_AI-based_Vessel_Segmentation/19688169}{https://figshare.com/articles/figure/FIVES\_A\_Fundus\_Image\_Dataset\_for\_AI-based\_Vessel\_Segmentation/19688169}} with 800 CFP images \cite{fives}, IOSTAR\footnote{\href{https://www.retinacheck.org/download-iostar-retinal-vessel-segmentation-dataset}{https://www.retinacheck.org/download-iostar-retinal-vessel-segmentation-dataset}} with 30 SLO images \cite{iostar}, PRIME-FP20\footnote{\href{https://ieee-dataport.org/open-access/prime-fp20-ultra-widefield-fundus-photography-vessel-segmentation-dataset}{https://ieee-dataport.org/open-access/prime-fp20-ultra-widefield-fundus-photography-vessel-segmentation-dataset}} with 15 UWF fundus images \cite{prime-fp20} and VAMPIRE\footnote{\href{https://vampire.computing.dundee.ac.uk/vesselseg.html}{https://vampire.computing.dundee.ac.uk/vesselseg.html}} with 8 FFA images \cite{vampire}. Each dataset is randomly partitioned into three disjoint subsets for training, validation, and test in a ratio of 7:1:2, respectively. An overview of the data is given in Tab. \ref{tab:dataset}.



\begin{table}[htb!]
    \caption{\textbf{Five public datasets used in this study}. }
    \label{tab:dataset}
    \renewcommand{\arraystretch}{1}
    \centerline{
        \scalebox{0.9}{
            \begin{tabular}{@{}l |l| r|r|r|r@{}}
               \hline
                 \textbf{Dataset} & \textbf{Modality} &\textbf{Total} & \textbf{Training}& \textbf{Validation} & \textbf{Test} \\
                \hline
    ROSSA \cite{octa-frnet} & OCTA   &918    & 642 & 92 & 184 \\
    \hline
    FIVES \cite{fives}      & CFP    & 800    & 560 & 80 & 160 \\
    \hline
    IOSTAR \cite{iostar}   & SLO    & 30     & 21  & 3  & 6   \\
    \hline
    PRIME-FP20 \cite{prime-fp20}  &  UWF   & 15     & 10  & 2  & 3   \\
    \hline
    VAMPIRE \cite{vampire}        &  FFA  &  8     & 5   & 1  & 2   \\
          
              \hline
            \end{tabular}
        }
    }
\end{table}


\textbf{Baseline Methods}. 
For a comprehensive comparison, we consider the following eight baseline methods: \\
$\bullet$ \texttt{Narrow-domain}: Five \texttt{R2AU-Net}s, separately trained per domain. \\
$\bullet$ \texttt{Narrow-domain-FT}: First training \texttt{R2AU-Net} on FIVES, \ie viewing CFP as the source domain, and then fine-tuning the model separately for each of the other four domains. \\
$\bullet$ \texttt{Broad-domain}: One \texttt{R2AU-Net} trained on the joint five-domain training set.  \\
$\bullet$ \texttt{prompt-DA} \cite{promptda}. Re-purposing \texttt{prompt-DA} for the BDRVS task, where CFP is treated as the source domain and the other four domains are regarded as the target domains. \\
$\bullet$ \texttt{prompt-SDA}. Note that \texttt{prompt-DA} treats target-domain training samples \emph{unlabeled}. For a more fair comparison, we extend \texttt{prompt-DA} to supervised domain adaptation \cite{cdcl} to learn from the labeled target-domain data.\\
$\bullet$ \texttt{MedSAM} \cite{medsam}: MedSAM, a state-of-the-art Vision Transformer based medical image segmentation network, fine-tuned on our broad-domain training data.\\
$\bullet$ \texttt{MedSAM-VPT}. Adapting MedSAM by visual prompt tuning (VPT) \cite{eccv22vpt}. \\
$\bullet$ \texttt{CVP} \cite{nips23cvp}: Prepending a domain-specific $3\times3$ conv. layer to the first encoder block of \texttt{R2AU-Net}.



\textbf{Implementation Details}. 
For a fair comparison, all  experiments are implemented as follows, unless otherwise specified.
The segmentation network is R2AU-Net \cite{r2attunet}, with its encoder  replaced by a pruned~\cite{prun} version of EfficientNet-B3 \cite{efficientnet}. 
All images are resized to 512$\times$512. 
Subject to our computing power (4 NVIDIA RTX 3090 GPUs), a mini batch has 5 images, one per domain. 
The network is trained to miniminze the mean of the BCE and Dice losses.
The optimizer is SGD with initial learning rate of 1e-3, momentum of 0.95 and weight decay of 1e-4. Learning rate is adjusted according to cosine annealing strategy \cite{loshchilov2016sgdr}.
Validation is performed every epoch. Early stop occurs if there is no increase in performance within 10 successive validations.
 PyTorch 1.13.1 is used. 


\textbf{Performance Metrics}. 
We primarily report pixel-level average precision (AP), more discriminative than Area under the ROC Curve (AUC) score. 

\begin{table}[htb!]
    \caption{\textbf{SOTA for BD-RVS}. Methods sorted  in ascending order by their mean AP on the five test sets.  Two naive baselines, \ie \texttt{Narrow-domain} with five domain-specific \texttt{R2AU-Net}s and \texttt{Broad-domain} with one \texttt{R2AU-Net}, are marked out in color. Compared with the best-performing baseline (\texttt{prompt-SDA}),  \texttt{DCP} is smaller and better.}
    \label{tab:sota}
    \renewcommand{\arraystretch}{1.1}
    \centerline{
        \scalebox{0.65}{
            \begin{tabular}{@{}l|r|r|r|r|r|r|r@{}}
               \hline
                \textbf{Method} & \textbf{Params}(M) &
                \textbf{Mean} & \textbf{ROSSA} & \textbf{FIVES} & \textbf{IOSTAR}& \textbf{PRIME-FP20} & \textbf{VAMPIRE} \\
                  \hline
                \texttt{prompt-DA}       & 16.9 & 0.3879 & 0.0553 & 0.8698 & 0.8071 & 0.1649 & 0.0425 \\
                \hline

                 \texttt{MedSAM-VPT}      & 91.7 & 0.5621 & 0.8042 & 0.7804 & 0.7528 & 0.1951 & 0.2779 \\
                \hline
                \texttt{MedSAM}   & 91.3 & 0.5692 & 0.8059 & 0.7923 & 0.7597 & 0.2041 & 0.2838 \\
                \hline
              
                \rowcolor{green!20} \texttt{Narrow-domain}  & 42.7 & 0.6670 & 0.8561 & 0.8867 & 0.8578 & 0.3720 & 0.3623 \\
                \hline
              
               
                \rowcolor{red!20} \texttt{Broad-domain}  & 8.5  & 0.6717 & 0.8386 & 0.8847 & 0.8559 & 0.4145 & 0.3646 \\
                \hline

                   \texttt{CVP}            & 8.5  & 0.6764 & 0.8417 & 0.8614 & 0.8362 & 0.3926 & 0.4502 \\
                \hline

                  \texttt{Narrow-domain-FT}     & 42.7 & 0.6790 & 0.8500 & 0.8867 & 0.8568 & 0.3916 & 0.4099 \\
                \hline

                
                \texttt{prompt-SDA}           & 16.9 & 0.6820 & 0.8452 & 0.8834 & 0.8466 & 0.4174 & 0.4175 \\
                \hline
                Proposed \texttt{DCP}      & 12.5 & \textbf{0.7037} & \textbf{0.8587} & \textbf{0.8928} & \textbf{0.8598} & \textbf{0.4298} & \textbf{0.4776}   \\  
          
              \hline
            \end{tabular}
        }
    }
\end{table}

\subsection{Comparison with SOTA}

As shown in Tab. \ref{tab:sota}, the \texttt{Broad-domain} method, despite its simplicity, is ranked at 4/8 among the eight baselines. In particular, it outperforms \texttt{Narrow-domain} on PRIME-FP20 and VAMPIRE, both of which have relatively limited training data. This result suggests that learning from the domain-mixed data is beneficial for the resource-limited domains. This is further confirmed by the better performance of \texttt{Narrow-domain-FT} against \texttt{Broad-domain} on VAMPIRE, obtained at the cost of using multiple models.
Compared with the best-performing baseline, \ie \texttt{prompt-SDA}, the proposed \texttt{DCP} is more accurate (0.7037 \emph{vs}. 0.6820 in AP), yet  smaller  (12.5M \emph{vs}. 16.9M in parameters).

\begin{figure}[htb!]
    \centerline{
    \includegraphics[width=0.45\textwidth]{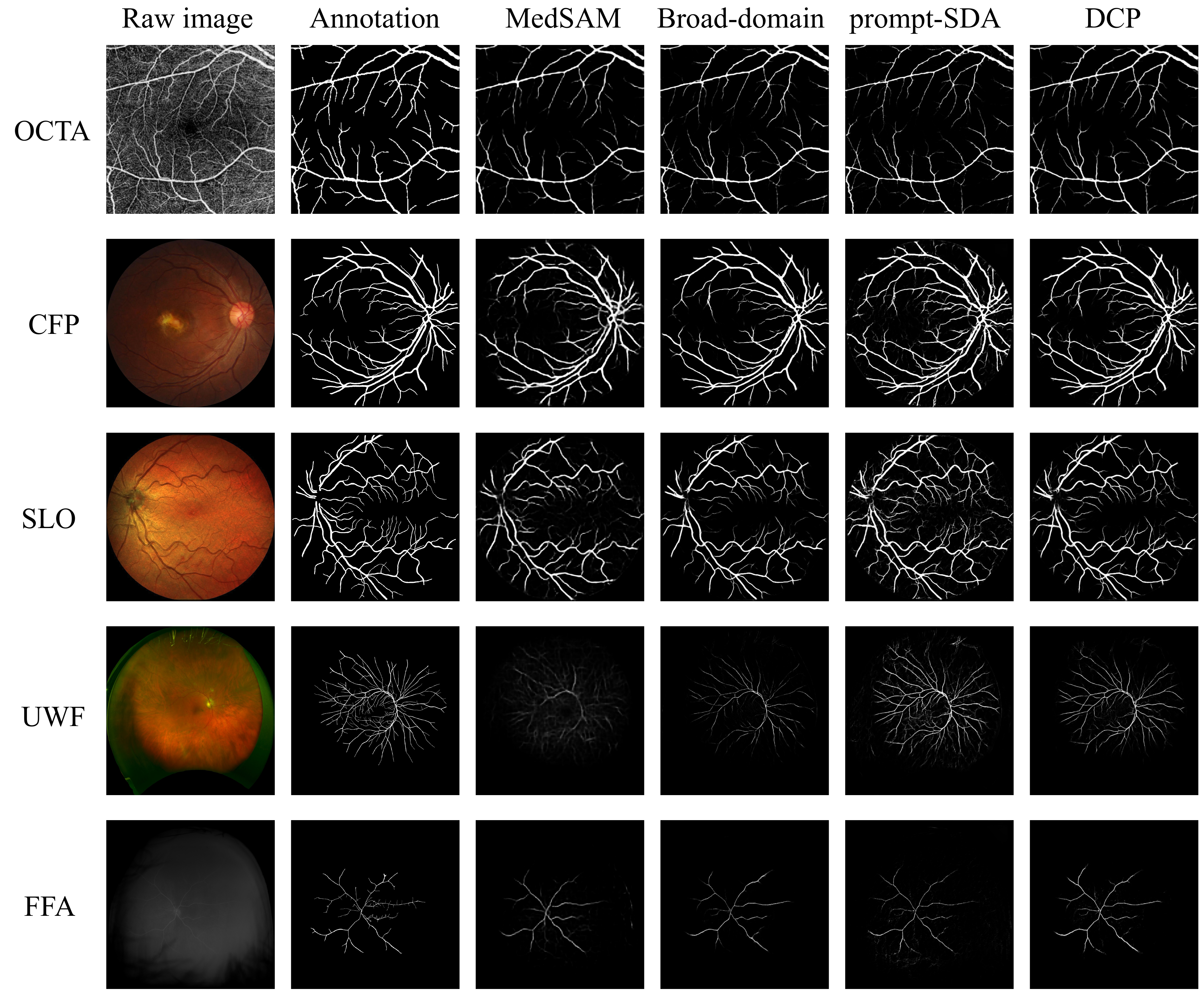}}
    \caption{\textbf{Qualitative results}. The efficacy of \texttt{DCP} is primarily manifested in segmenting capillaries.
    Best viewed digitally. }
    \label{fig:showcase}
    \vspace{-0.2cm}
\end{figure}

In addition, we report the AUC scores in Tab. \ref{tab:auc}. 
The AUC-based ranking of the different methods is largely consistent with its AP counterpart, except that \texttt{CVP} becomes the best baseline. Our \texttt{DCP} again surpasses the baselines. Some qualitative results are shown in Fig. \ref{fig:showcase}. 

\begin{table}[t!]
    \caption{\textbf{Performance measured by AUC}.}
    \label{tab:auc}
    \renewcommand{\arraystretch}{1.1}
    \centerline{
        \scalebox{0.65}{
            \begin{tabular}{@{}l|r|r|r|r|r|r|r@{}}
               \hline
                \textbf{Method} &
                \textbf{Mean} & \textbf{ROSSA} & \textbf{FIVES} & \textbf{IOSTAR}& \textbf{PRIME-FP20} & \textbf{VAMPIRE} \\
                \hline
                \texttt{prompt-DA}  & 0.8167 & 0.4783 & 0.9782 & 0.9784 & 0.8769 & 0.7717\\
                \hline

                \texttt{MedSAM-VPT} & 0.9438 & 0.9845 & 0.9643 & 0.9644 & 0.8876 & 0.9183\\
                \hline
                \texttt{MedSAM}   & 0.9451 & 0.9849 & 0.9675 & 0.9666 & 0.8897 & 0.9166 \\
                \hline
              
                \rowcolor{green!20} \texttt{Narrow-domain}  & 0.9537 & 0.9886 & 0.9837 & 0.9791 & 0.9204 & 0.8968 \\
                \hline
              
                \rowcolor{red!20} \texttt{Broad-domain}  & 0.9664 & 0.9873 & 0.9838 & 0.9857 & 0.9408 & 0.9345  \\
                \hline

                \texttt{Narrow-domain-FT} &  0.9664 & 0.9886 & 0.9837 & 0.9820 & 0.9284 & 0.9492\\
                \hline
                
                \texttt{prompt-SDA} &  0.9680 & 0.9873 & 0.9823 & 0.9825 & 0.9437 & 0.9444 \\
                \hline
                \texttt{CVP} & 0.9689 & 0.9873 & 0.9804 & 0.9823 & 0.9422 & 0.9521\\
                \hline
                Proposed \texttt{DCP} & \textbf{0.9739} & \textbf{0.9887} & \textbf{0.9862} & \textbf{0.9862} & \textbf{0.9537} & \textbf{0.9549}   \\  
          
              \hline
            \end{tabular}
        }
    }
\end{table}


\subsection{Ablation study}

Tab.~\ref{tab:abl} shows the performance of \texttt{DCP} with varied setups. Removing feature partition / position-wise (pos.) prompting / channel-wise (cha.) prompting consistently results in performance loss. The necessity of these component is thus justified. 
We also check if similar improvement can be obtained by adding MHSAs for position-wise / channel-wise feature interaction \emph{without} using any prompt, see ``\emph{w/o} prompt'' in Tab. \ref{tab:abl}. Its lower performance than ``Full'' verifies the importance of the prompts. 

 \begin{table}[htb!]
    \caption{\textbf{Ablation study}. 
    The performance gap of not using a specific component to the full setup reflects the importance of that component. Metric: AP.}
    \label{tab:abl}
    \renewcommand{\arraystretch}{1.2}
    \centerline{
        \scalebox{0.65}{
            \begin{tabular}{@{}l |r| l|r|r|r|r|r @{}}
               \hline
                \textbf{Setup} 
                & \textbf{Params}(M) & \textbf{Mean} & \textbf{ROSSA} & \textbf{FIVES} & \textbf{IOSTAR}& \textbf{PRIME-FP20} & \textbf{VAMPIRE} \\
                \hline
                \emph{w/o} partition        & 16.3 & 0.6781(\textcolor{red}{$\downarrow$3.6\%}) & 0.8480 & 0.8839 & 0.8434 & 0.4155 & 0.3997 \\
                \hline
                \emph{w/o} prompt           & 12.2 & 0.6833(\textcolor{red}{$\downarrow$2.9\%}) & 0.8554 & 0.8775 & 0.8527 & 0.4066 & 0.4241 \\
                \hline
                \emph{w/o} pos. prompting  & 9.1 & 0.6896(\textcolor{red}{$\downarrow$2.0\%}) & 0.8394 & 0.8908 & 0.8489 & 0.4205 & 0.4483 \\
                \hline
                \emph{w/o} cha. prompting   & 10.8 & 0.6915(\textcolor{red}{$\downarrow$1.7\%}) & \textbf{0.8610} & 0.8828 & 0.8520 & 0.4041 & 0.4577 \\
                \hline
                Full                        & 12.5 & \textbf{0.7037} & 0.8587 & \textbf{0.8928} & \textbf{0.8598} & \textbf{0.4298} & \textbf{0.4776}   \\  
            \hline
            \end{tabular}
        }
    }
\end{table}




\section{Conclusions}

For broad-domain retinal vessel segmentation (BD-RVS), we propose dual convolutional prompting (\texttt{DCP}), a plug-in module effectively turning an R2AU-Net based vessel segmentation network to a unified model that works for CFP, OCTA, SLO, UWF and FFA. Experiments on a broad-domain dataset verifies the effectiveness of the proposed method. Both position-wise and channel-wise prompting are useful. Localized prompting also matters. We believe our study of BD-RVS has opened new opportunities for the long-studied RVS problem. 

\balance
\bibliographystyle{IEEEtran}
\bibliography{ref}

\end{document}